\documentclass[journal]{IEEEtran}

\usepackage{etoolbox}
\makeatletter
\patchcmd{\@makecaption}
  {\scshape}
  {}
  {}
  {}
\makeatother
\ifCLASSINFOpdf
\else
\fi
\usepackage{amsmath}
\usepackage{graphicx}
\usepackage{subfigure}
\usepackage{float}
\usepackage{multirow}
\usepackage{booktabs}
\usepackage{pgffor,tikz}
\usepackage{booktabs}
\usepackage{array, caption, threeparttable}
\usepackage[font=normalsize,labelfont=normalsize,labelsep=colon]{caption}
\usepackage{hyperref}
\hypersetup{hypertex=true}
\begin{document}
%
\title{Neural Image Beauty Predictor\\ Based on Bradley-Terry Model}
%
%
%

\author{Shiyu~Li,~\IEEEmembership{}
        Hao~Ma,~\IEEEmembership{}
        and~Xiangyu~Hu ~\IEEEmembership{}
        
\thanks{
S. Li, Hao Ma, X. Hu, Department
of Engineering Physics and computation, Technical University of Munich, 
Garching, 85748 Germany Correspondence to: X. Hu  e-mail: xiangyu.hu@tum.de}
}

\maketitle

\begin{abstract}
Image beauty assessment is an important subject of computer vision. Therefore, building a model to mimic the image beauty assessment becomes an important task.  To better imitate the behaviours of the human visual system (HVS), a complete survey about images of different categories should be implemented. This work focuses on image beauty assessment. In this study, the pairwise evaluation method was used, which is based on the Bradley-Terry model. We believe that this method is more accurate than other image rating methods within an image group. Additionally, Convolution neural network (CNN), which is fit for image quality assessment, is used in this work. The first part of this study is a survey about the image beauty comparison of different images. The Bradley-Terry model is used for the calculated scores, which are the target of CNN model. The second part of this work focuses on the results of the image beauty prediction, including landscape images, architecture images and portrait images. The models are pretrained by the AVA dataset to improve the performance later. Then, the CNN model is trained with the surveyed images and corresponding scores. Furthermore, this work compares the results of four CNN base networks, i.e., Alex net, VGG net, Squeeze net and LSiM net, as discussed in literature. In the end, the model is evaluated by the accuracy in pairs, correlation coefficient and relative error calculated by survey results. Satisfactory results are achieved by our proposed methods with about 70 percent accuracy in pairs. Our work sheds more light on the novel image beauty assessment method. While more studies should be conducted, this method is a promising step.

\end{abstract}


%
\IEEEpeerreviewmaketitle

\section{Introduction}
%
%
%
%
\IEEEPARstart{C}{omputer}  vision studies the behavior of human visual system (HVS). The aim of computer vision could be to mimic the decision of HVS by computer programs. HVS can be used as an efficient tool to evaluate images from portraits to landscapes. Several studies show that non-professionals can provide reliable results, which are applied in physical \cite{larkin1980expert} or aesthetic problems \cite{murray2012ava}. If neural network models are trained based on HVS, the problems of image assessment can be solved.

\subsection{Motivation}
Images are useful media to transfer information. They are not only widely used to present results in natural sciences, but also used to convey emotion or show beauty. However, the criteria to evaluate an image are problematic. High-quality images need to be chosen from the image database, but the evaluation does not have references or clear standards. In terms of aesthetic assessment, to pick out the most beautiful images is a great challenge.

There are many methods to judge the quality of an image. However, these methods do not always converge to the same conclusion, while HVS can always make a clear decision. Individuals will have different preferences, but a large group will show a clear conclusion for any case.

The benefits of automatic evaluation, where the neural network models can replace the HVS, include lower costs and less survey resources. Lower costs can be reached since no surveys are required. Considering the resources required by surveys implemented by humans, deep learning methods can work with higher efficiency.


\subsection{Objective and our contribution}
The aim of this work is to develop a convolutional neural network (CNN) to replace HVS for image assessment. Image assessment includes detection or quality evaluation of different images, which covers many fields of applications, such as aesthetic evaluation. However, the assessment of images cannot be easily implemented in consideration of different research subjects. HVS, in comparison with other methods, has great advantages in accuracy and robustness. Accuracy means that the surveyed people will give a comprehensive assessment with conclusive results. Robustness means that the results of surveys do not change much with an increasing number of survey samples. The main problem of crowdsourcing by HVS as mentioned is time-consumption. It would be impossible to perform an image assessment manually for every problem. A model based on the results of HVS could be a good alternative.

In this work, a model which predicts the survey results of HVS will be given and the neural models should not only predict the priority of most raters, but also the winning probability of each answer. The evaluation based on HVS should be realized by the Bradly-Terry model to discover the real preference of raters. The simple evaluation of one single image is not enough to compare images since the comparison may vary with different subjects and different crowds of people. The comparison in pairs, in contrary, is more convincing due to more consideration in each pair among each image group. Therefore, the objective is to build a model in a competitive way and to train a neural network based on this model. Finally, the trained CNN model should give a reasonable prediction about the preference of HVS with enough accuracy.

Our source codes, images and survey results are available at \url{https://github.com/lishiyu0088/Neural_Bradley-Terry}.

\section{Related Work and Preliminary Work}
\subsection{Bradley-Terry model}
In a case where the participants in a group are asked to be repeatedly compared with one another in pairs, Bradley and Terry (1952) suggested the model \cite{bradley1952rank}
\begin{equation}
p_{ij}=\frac{\gamma_i}{\gamma_i+\gamma_j}
\end{equation}
where $p_{ij}$ is the probability that raters choose sample $i$ over sample $j$, $\gamma_i$ is a positive-valued parameter associated with sample $i$, for each of the comparisons between samples $i$ and sample $j$. As a common example, consider the samples to be different image candidates, where $\gamma_i$ represents the overall evaluation of sample $i$.

After collecting the votes for a set of pairwise comparisons among $N$ samples (i.e., images) from a user study of different images, we transform the participants’ pairwise answers into a set of scores, $s=\left\{s_1,s_2,\ldots,s_N\right\}$, using the Bradley-Terry model. Then
$\gamma_i$ can be expressed by an exponential function of scores:
\begin{equation}
\gamma_i=e^{s_i}
\end{equation}
The formula above can be reformulated as:
\begin{equation}
p_{ij}=\frac{e^{s_i}}{e^{s_i}+e^{s_j}}=\frac{e^{s_i-s_j}}{1+e^{s_i-s_j}}
\end{equation}

Consider an extension of the Bradley-Terry model to comparisons concerning more than three samples, where such a comparison between samples are supposed to be ranked from best to worst as a result. The scores should be based on the results of comparison in pairs and give an overall image of every sample.  A complete survey of models for this type is proposed by Marden (2019) \cite{marden2019analyzing}.

Suppose that there are $m$ samples, labeled $1$ through $m$, in the assumed group. For $A\subset\left\{1,\ \ldots,\ k\right\}$ with $k\le m$, suppose that the samples indexed by A are to be evaluated. To represent the order of priority, $\rightarrow$ denotes the relation "is ranked higher than" and $S_k$ denotes the group of k elements sorted from the best to the worst. Given A and some $\pi\in S_k$, the probability of the event $\pi(1)\rightarrow\cdots\pi(k)$ is:
\begin{equation}
P_A\left[\pi(1)\rightarrow\cdots\pi(k)\right]=\prod_{i=1}^{k}{\frac{\gamma_\pi(i)}{\gamma_\pi\left(i\right)+\cdots\gamma_\pi\left(k\right)}.}
\end{equation}
The winning probabilities of multiple comparisons can be calculated by the conditional probabilities formulated as follows \cite{hunter2004mm}:
\begin{equation}
P_B\left(i\ wins\right)=P_A(i\ wins)P_B(anything\ from\ A\ wins)
\end{equation}
for all $i \in A\subset B$.

The winning probability of A can be calculated by the ranking process as first choosing a winner and then adding all the other following samples, which is written as follows:
\begin{equation}
\begin{aligned}
&P_A\left(i\ wins\right)=\sum_{\pi:\pi\left(1\right)=i}{P_A\left[\pi\left(1\right)\rightarrow\cdots\rightarrow\pi\left(k\right)\right]}\\
=&\sum_{\pi:\pi\left(1\right)=i}\frac{\gamma_i}{\gamma_1+\cdots+\gamma_k}\prod_{j=2}^{k}\frac{\gamma_\pi(j)}{\gamma_\pi\left(j\right)+\gamma_\pi\left(j+1\right)\cdots\gamma_\pi\left(k\right)}\\
=&\frac{\gamma_i}{\gamma_1+\cdots+\gamma_k}
\end{aligned}
\end{equation}

\subsection{Crowdsourcing Survey}
Crowdsourcing is s a “problem-solving approach that tapes the 
knowledge, energy, and creativity of a global, online community" \cite{parvanta2013crowdsourcing}. In this work, crowdsourcing is used by gathering the votes for image evaluation. HVS plays an important role in this process, since every participant uses its visual system for voting. Crowdsourcing methods can be used in image evaluation for many fields of research, since it has also demonstrated its ability to support science projects via crowdsourcing, e.g., for experimental behavioral research \cite{crump2013evaluating}, scoring chemistry images \cite{irshad2017crowdsourcing}, and astronomical discovery.

Um et al. uses HVS for the evaluation of image similarity between fluid simulation images \cite{um2021spot}. The participants are asked to choose a more similar image compared with a reference image. Images are compared in a group of pairs. They get a converged result with 50 participants. Their methodology gives a clear answer to many unsolved image evaluation problems. It is proved to be more reliable than other tradition similarity metrics.

\subsection{Image assessment models}
Image assessment  includes the evaluation of image similarity, image quality assessment (IQA) or range of distortion including blocking, blur or noise \cite{brandao2010no, damera2000image, ou2010perceptual, seshadrinathan2009motion, sheikh2005no, tong2005learning}. CNN is capable of processing images with different levels of features. Therefore, it is widely used in image assessment. Zhang et al. developed a perceptual metric for similarity metrics. In the framework of LPIPS \cite{zhang2018unreasonable}, feature maps from different neural layers serve as the inputs for the results. They also compared framework of Alex net, Squeeze net and VGG net. Kohl et al. developed a CNN similarity model especially for fluid simulation \cite{kohl2020learning}. During the training process, a loss function which combines a correlation loss term with a mean squared error is adopted to improve the accuracy of the learned metric.

Mittal et al. proposed an unsupervised, training free CNN model \cite{mittal2011blind}. It assumes that the distorted images have certain characteristics compared with natural images. The results correlated well with LIVE data base. Xue et al. made a IQA without human scores by applying Gaussian noises, Gaussian blur, distortion, or image compression on the original images \cite{xue2013learning}. FSIM is used as the criteria of image quality for the training process.

\subsection{Image aesthetic models}
Traditional machine learning approaches use extracted features to evaluate the image beauty \cite{datta2006studying, ke2006design}. These features can be composition, contrast or hues. Some of the image aesthetic models use Support-Vector Machine (SVM) classifier to predict image aesthetic quality \cite{bianco2018aesthetics, luo2011content}. Other proposed methods, like Fisher Vector \cite{marchesotti2011assessing, perronnin2007fisher, perronnin2010improving} and visual words \cite{su2011scenic} can also be used for aesthetic prediction. However, these approaches have their limitation and they are now outperformed by deep learning methods \cite{hii2017multigap, karayev2013recognizing,ma2017lamp, karpathy2015deep, tang2014blind}.

CNN is a powerful tool to predict image aesthetic quality, as it is  widely used in recent years. \cite{jin2016image}. Most training in the literature is based on the AVA dataset and the scores are used for a regression model. This model can predict the histogram prediction instead of classifying images by the mean score. Kong et al. built a photo aesthetic network, which considered the judgement of different individuals \cite{kong2016photo}. They built a new database, i.e. AADB, which recorded the ID of different viewers. They compare the images in pairs to include more information for the training. Besides regression loss, rank loss is also included. Talebi \& Milanfar make an improvement based on the previous work \cite{bianco2018aesthetics} \cite{mittal2011blind}. Their datasets include AVA, Tampere and LIVE. They used a cross validation method to check their model between different datasets. Their models have predictions with higher correlation \cite{talebi2018nima}. In terms of the evaluation for portrait images, Lienhard et al. proposed a database, which uses Human Faces Score (HFS) to evaluate them \cite{lienhard2015predict}. Bianco et al. use CNN for aesthetic prediction on portrait images, which include the factors of facial expressions, brightness, contrast as influences \cite{bianco2018aesthetics}. Their training data includes CUHKPQ, HFS, FAVA and Flickr data base. Their predicting results are evaluated by good classification rate (GCR) and Pearson’s correlation coefficient, which shows improvement compared with previous methods.

As the aesthetic evaluation of images should include global and local features, patches of different size are extracted as inputs for CNN model \cite{lu2014rapid}. Based on patches, Lu et al. used aggregated patches as an input of DMA-net for aesthetic prediction \cite{lu2015deep}. Sheng et al. used Attention-based Multi-Patch Aggregation for processing patches of images, which improve the performance \cite{sheng2018attention}. In addition to patches, Mai et al. proposed a composition-preserving deep network for photo aesthetics assessment \cite{mai2016composition}. Instead of fixing the size of receptive field, they fix the ouput dimension by adaptive spatial pooling layer (ASP), which is able to handle images with different sizes and aspect ratios.
\subsection{Dataset of Aesthetic Evaluation}
There are many available datasets for aesthetic evaluation in the literature. These datasets are based on a voting system from internet. The aesthetic evaluation can be represented by scores given by the audience. However, scores may not be accurate since the images are not in the same categories. In this section, part of current datasets will be introduced and compared with our dataset.
\subsubsection{A large-Scale Database for Aesthetic Visual Analysis (AVA)}
The AVA dataset \cite{murray2012ava} is the largest aesthetic dataset among the current literature. It contains about 255,000 images, which include all kinds of categories from landscapes to portraits. Each image has a large amount of useful information, which can be used for aesthetic analysis. Among them, the aesthetic scores are the most important, which are given by website visitors and photographers. These scores range from 1 to 10, which represents the worst to the best image. The number of votes for each score is counted. Therefore, the average score and deviation are valuable standards for evaluation. These images are ranked by index and image ID. These images belong to different challenge themes, which are labelled by challenge ID.

The shortcoming of this dataset is that different types of images are evaluated together, rather than categorized in small groups, like portrait, landscape. On the one hand, these images have different themes. There are different standards to evaluate them. On the other hand, they are evaluated by different groups of people. Therefore, photos in different groups may not be comparable. In some special case, some images have a high quality, but receive a low score because it does not correspond to the specific challenge topics. These images are not suitable for training CNN models.

\subsubsection{CUHKPQ}
The CUHKPQ dataset consists of 17,673 images in seven categories \cite{luo2011content}, i.e. animal, plant, static, architecture, landscape, human and night. The advantage of this dataset is that images from different categories can be analysed by different models, which improve the accuracy. The images are divided into groups of high and low quality. Unlike the AVA dataset, these images are evaluated by professional photographers instead of amateurs. Regional and global features are used for analysis of the aesthetic quality.

However, the CUHKPQ dataset only evaluates images qualitatively. It does not offer a score, but divide images simply into good and bad quality groups, which can be rather extreme and arbitrary. In many cases, the difference of images are quite small and the evaluation between them could be very difficult. This dataset is not able to represent the aesthetic standard accurately.
\subsubsection{Our dataset}
Our dataset combines the advantages of the AVA and the CUHKPQ dataset. It includes 900 images, which can be divided into three categories, i.e. portrait, architecture and landscape. Our dataset tries to compare the aesthetic quality of images of the same type, since different kinds of images are not comparable. For example, the quality of landscape cannot directly be compared with that of portrait. Another issue of this dataset tries to solve is to make the scores of different images ranked in the same range. A simple rating from 1 to 10 is sometimes ambiguous for the viewers, because viewers are not sure which standards each number represents.

Our dataset evaluates image aesthetic quality by the Bradley-Terry model. Only images from same categories are compared with each other. First of all, viewers should compare two images and choose the more attractive image. After voting, the scores can be calculated according to the model. This method should be more accurate, because the scores are given by comparison in pairs.

This dataset has fewer images compared with the AVA and the CUHKPQ dataset, since the surveys are limited in the number of participants. It aims to be more precise on the voting system instead of providing a overall image beauty estimation. Our dataset tries to find the characteristics of the HVS specific to the type of image, for instance portrait, architecture and landscape.
\section{Image beauty survey}
This section introduces the content, process, and results of the image beauty survey. The survey results are used for training of different CNN models.
\subsection{Survey content and process}
\begin{figure}
  \centering
  \includegraphics[scale=0.8]{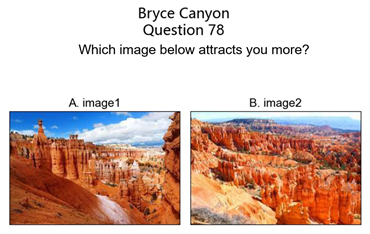}\\
  \caption{Survey question design of landscape images: All user studies in this paper were conducted with the design shown above. Each participant must make a choice between A and B per question.}
  \label{img_survey}
\end{figure}

In order to evaluate the data efficiently, we made the question as simple as possible. This helps reduce misunderstanding and guarantee the quality of the survey result. In the survey, the participants are asked to choose the more attractive image among the two images. Due to the difficulty in choosing the images in a group, the survey should be implemented in pairs, i.e., two-alternative forced choice, where the participants are given two options and must choose one of them. The comparison is made without a reference image. Fig. \ref{img_survey} shows the survey question design of landscape images. This survey uses a crowdsource method for evaluation. As the number of participants increases to more than 50, the result turns out to be stable. The design of survey for portrait and architecture images follows the same pattern.

\subsection{Survey result}
This part discusses parts of the survey result for image beauty assessment for landscapes, portrait and architecture. The calculation process and results  are shown in detail in Appendix B and C. 
Fig. \ref{yosemite} shows a result of one landscape sample, which evaluates the quality of different images from Yosemite National Park of America. Image 4 has the highest score 1.106, followed by image 6 with score 1.000. Image 1 has the lowest score 0.339. The results of this evaluation are associated with color variety.
\begin{figure}[h]
\centering
\subfigure[Survey images]{
  \foreach \j in {1,2,...,6}  {
    \includegraphics[height=0.08\linewidth]{Appendix_image/Landscape/27/\j.jpg} 
  }
  }
\centering
\subfigure[Survey results with exponential scores]{
\includegraphics[scale=0.5]{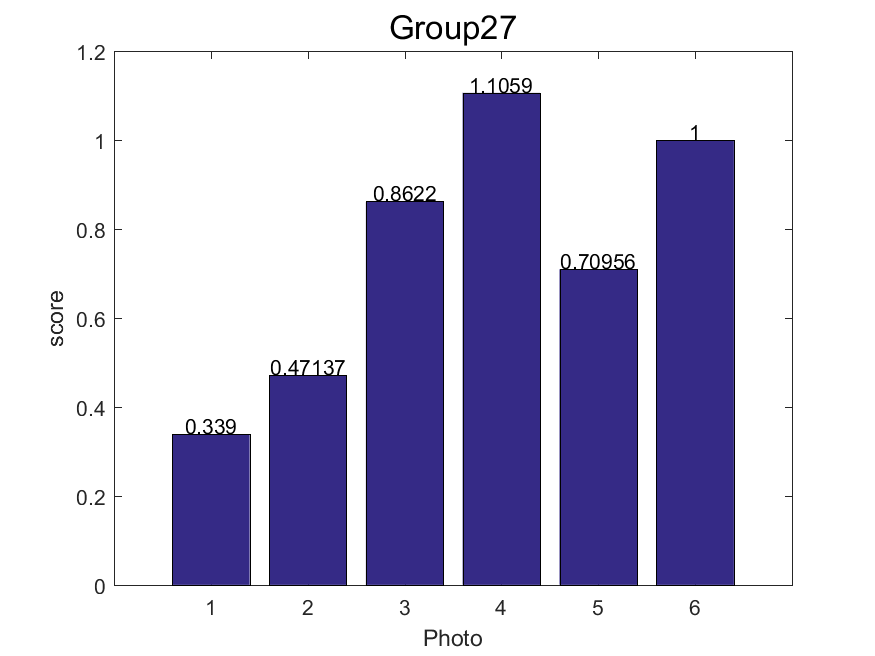}}
   \caption{Survey images and result of Yosemite.}
   \label{yosemite}
\end{figure}

\begin{figure}[h]
\centering
\subfigure[Survey images]{
  \foreach \j in {1,2,...,5}  {
    \includegraphics[height=0.17\linewidth]{Appendix_image/Face/52/\j.jpg} 
  }
  }
\centering
\subfigure[Survey results with exponential scores]{
\includegraphics[scale=0.5]{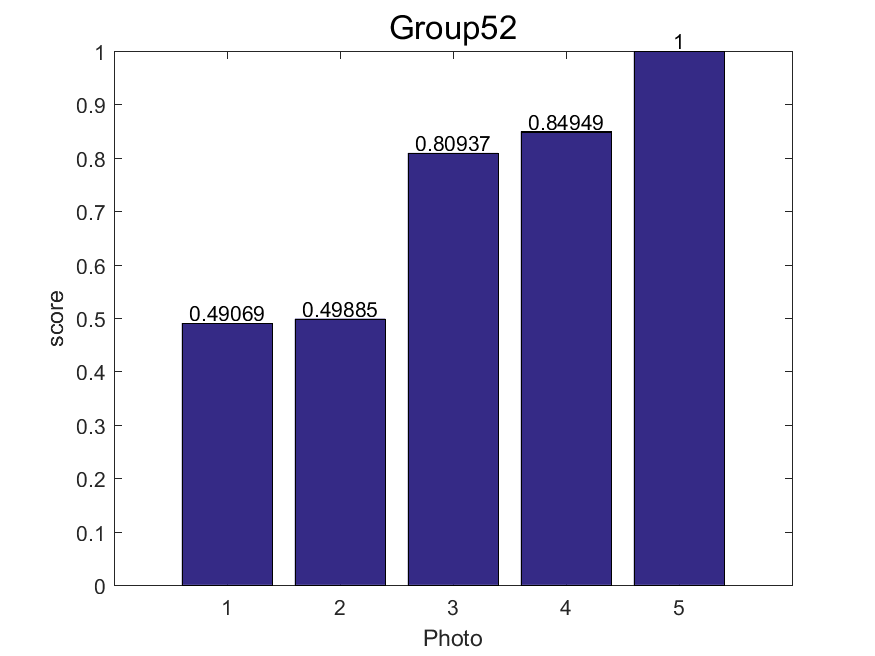}}
   \caption{Images of Human face: young woman with 15-30 years old and survey result.}
   \label{woman}
\end{figure}

Fig. \ref{woman} shows images of young women at the age of 15-30. Image 5 of a smiling face with red hair has the highest score 1.000. Then image 4 has the second rank with 0.850. Image 1 has the lowest score 0.491. The scores from image 1-5 increase continuously.

\begin{figure}[h]
\centering
\subfigure[Survey images]{
  \foreach \j in {1,2,...,6}  {
    \includegraphics[height=0.09\linewidth]{Appendix_image/Buildings/21/\j.jpg} 
  }
  }
\centering
\subfigure[Survey results with exponential scores]{
\includegraphics[scale=0.8]{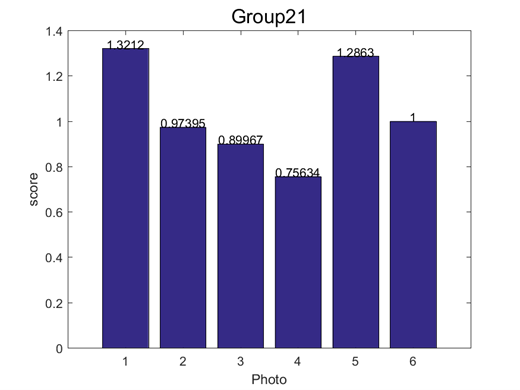}}
   \caption{Survey result of group 21 for building images: Statue of Liberty.}
   \label{statue}
\end{figure}

Fig. \ref{statue} shows one result of an architecture survey, which evaluates the image quality of the Statue of Liberty. In the survey, image 1 has the highest score with the score 1.321, followed by image 5 with the score 1.286. These two images have higher contrast and more details than the other images. Image 4 has the lowest score 0.756. This may be due to the monotonous background.

\section{Constructing a CNN based images beauty assessment model}
\subsection{Division of dataset and data loader}
\begin{figure}
  \centering
  \includegraphics[scale=0.6]{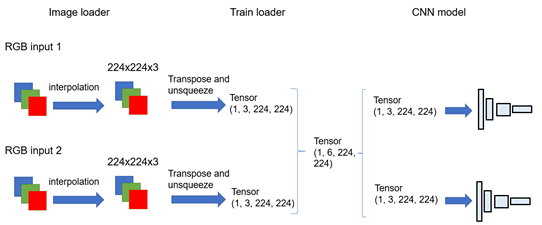}\\
  \caption{Train loader of image pairs.}
  \label{loader}
\end{figure}

Before the training process begins, the data set should be divided into training, validation, and testing parts. Landscape, portrait, and architecture images are trained separately. About 20\% of the data is used for validation and 20\% of the data is used for test.
During the loading process, each possible combination of two images in a group are selected and then merged into one tensor in the shape of $(1, 6, 224, 224)$, as shown in Fig. \ref{loader}. In the training process, the loaded image groups are shuffled into random order to train the CNN model more evenly.  Then the combined tensor is split into two images again, each in the tensor shape of $(1, 3, 224, 224)$, and will be fed through the model.

\subsection{Pretrained CNN based on AVA dataset and Results}
In this work, AVA dataset is used to pretrain the network. As AVA has a great amount of images, CNN can fully generate feature maps, and recognize as many image characteristics  as possible. There are four kinds of base nets, followed by fully connected layers. In this process, the average voting scores are used as the target for every input image. In contrast to other literature, we will not focus on deviation of scores. Therefore, our model only predicts the average voting score instead of the distribution.

\begin{figure}
  \centering
  \includegraphics[scale=0.3]{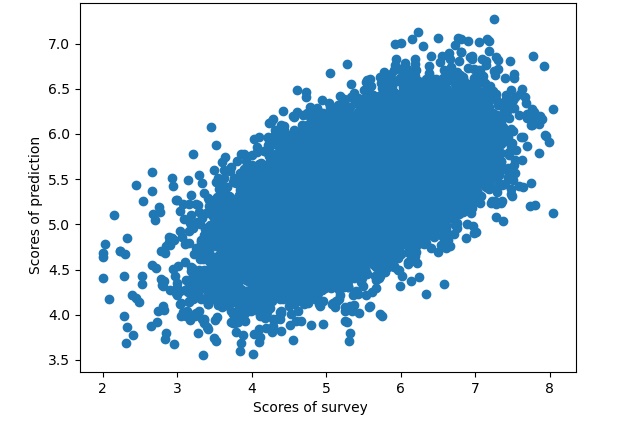}\\
  \caption{Scatter plot of the validation data for squeeze base net: In this figure, x-axis represents the scores of survey, while y axis represents the predicted scores by squeeze base net. A linear correlation can be observed from this plot.}
  \label{scatter}
\end{figure}

The training process takes 25 epochs until the training loss is stable and the validation correlation reaches a plateau. The VGG net and Squeeze net take longer than the Alex base net and the LSiM net, since the former nets have more parameters. Fig. \ref{scatter} presents the scatter plot of the validation data. A linear correlation between the voting score and predicted score can be observed. Most of the voting scores concentrate around the score 5, while only few images get lower or higher scores.
Table I shows the prediction results of four different base nets. VGG has the best result with a correlation coefficient of 0.650, which is better than the correlation coefficient found in literature \cite{kong2016photo}\cite{talebi2018nima}.

\begin{table}[htbp]
\centering
\caption{Spearman's Rank Correlation coefficient (SRCC) of results for different pretrained CNN models based on AVA dataset: Bold and underlined value shows the best performance of each dataset.}
\begin{tabular}{ccc}
\hline
Models & Correlation of validation & Correlation of test \\ \hline
Kong et. al \cite{kong2016photo} & - & 0.558 \\
NIMA (MobileNet) \cite{talebi2018nima} & - & 0.510 \\
NIMA (VGG16) \cite{talebi2018nima} & - & 0.592 \\
NIMA (Inception-v2) \cite{talebi2018nima} & - & 0.612 \\
Alex base net &0.532& 0.526 \\
VGG base net &\underline{\textbf{0.646}}& \underline{\textbf{0.650}} \\
Squeeze base net &0.576& 0.597 \\
LSiM base net &0.400& 0.404 \\ \hline
\end{tabular}
\end{table}

These pretrained base nets serve as initial values for the training model used later. The AVA data set, which includes various categories, is large enough for the pretraining. These models are useful for the fine-tuning process in specific image evaluation. It improves the prediction result of our dataset.
\subsection{Training architecture of our dataset}
\begin{figure}
  \centering
  \includegraphics[scale=0.55]{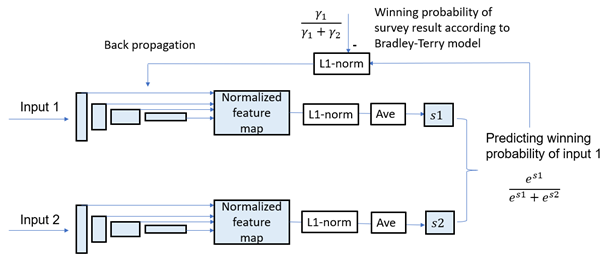}\\
  \caption{Architecture of CNN model.}
  \label{cnn}
\end{figure}

Figure \ref{cnn} shows the architecture of the model. First, normalized feature maps are extracted from the CNN base net. Then the average values $s_1$ and $s_2$ are calculated from the L1-norm of the normalized feature map. The predicting winning probability is calculated from the $s_1$ and $s_2$ according to the Softmax function. After that, a backpropagation is implemented  by introducing the aesthetic scores from survey results.

The loss function of the CNN model is defined as the $L_1$ loss of the difference between the predicted winning probability and survey result, which is shown as follows:
\begin{equation}
L=\left|\frac{\gamma_1}{\gamma_1+\gamma_2}-\frac{e^{s_1}}{e^{s_1}+e^{s_2}}\right|
\end{equation}

\section{Results}
This part discusses the validation and test process of CNN model for   landscape, architecture and portrait images. Different CNN models are compared with survey results. These models are trained individually since images from different categories have their own special properties. The goal of our model is to train and evaluate the images in three different classes, which is only seldom considered by other literature. The results introduced in this part are based on our dataset. Before the presentation of our result, some of the details used in our models are explained below.

The CNN models used in this paper are implemented in Pytorch. Especially the CNN models pretrained by the AVA dataset as discussed before are used in this part to improve the performance of models. In the pretrained CNNs, all the values are initialized by corresponding models. During the training process, the first three layers in the base net are frozen. The last two layers and the rest of the linear layers are fine-tuned by the image data. The results of pretrained networks by the AVA dataset and networks of initial values will be compared with each other. Adams optimizer is used to find the solution for this problem and initial learning rate is set to $5 \times 10^{-5}$. As the epochs increase, learning rate is adjusted with decay to have better convergence.
\subsection{Evaluation of performance}
To compare the performance of different models, several parameters are used to evaluate the CNN models, i.e. accuracy of pairs, Pearson's correlation coefficient (PCC), Spearman's rank correlation (SRCC) and relative error of prediction. These parameters, which analyse results in different perspectives, are defined and described as follows.

First, as the images are loaded in groups, accuracy of pairs can be used to evaluate the quality of prediction. A group of images are compared with each other in pairs. If the result of the CNN model is in accordance with the result of the survey, which means one image has higher score than another both in the survey and the prediction results, then this pair is regarded to be labelled correctly. Accuracy of pairs are defined as the ratio of correctly labelled pairs to the number of all pairs in one group.

Pearson’s correlation coefficient (PCC) is commonly represented by $r_{xy}$ and may be referred to as the sample correlation coefficient or the sample Pearson correlation coefficient \cite{pearson1895notes}. The formula is obtained by dividing the multiplication of covariances by individual variances based on the sample. Given paired data $\left\{(x_1,y_1),\cdots,(x_n,y_n)\right\}$ consisting of n pairs, $r_{xy}$ is defined as follow:
\begin{equation}
r_{xy}=\frac{\sum_{i=0}^{n}{(x_i-\bar{x})(y_i-\bar{y})}}{\sqrt{\sum_{i=0}^{n}{(x_i-\bar{x})}^2}\sqrt{\sum_{i=0}^{n}{(y_i-\bar{y})}^2}}
\end{equation}
where n is the sample size, $x_i,\ y_i$ are the survey scores and predicted scores indexed with i and $\bar{x}=\frac{1}{n}\sum_{i=1}^{n}x_i$ (the sample mean); and $\bar{y}=\frac{1}{n}\sum_{i=1}^{n}y_i$.

Spearman’s correlation coefficient (SRCC) does not consider the absolute value, but it only considers the ranking order \cite{spearman1961proof}. The Spearman correlation coefficient is defined as the Pearson correlation coefficient between the rank variables. For the sample of the size n, the n raw scores $x_i,\ y_i$ are converted to ranks ${rg}_{x_i},\ {rg}_{y_i}$, and $r_s$ is computed as
\begin{equation}
r_s=\rho_{{rg}_X,{rg}_Y}=\frac{cov(rgX,rgY)}{\sigma_{{rg}_X}\sigma_{rgY}}
\end{equation}

Finally, the results of the CNN model will be quantitatively evaluated by relative errors. The relative error can be calculated by the difference between predicting winning probability $P_{CNN}$ of the CNN model and the winning probability $P_{survey}$ calculated from the survey, which is defined as follows:
\begin{equation}
e_r=\frac{\left|P_{CNN}-P_{survey}\right|}{P_{CNN}}
\end{equation}
\subsection{Results of Landscape Images Assessment}
\begin{figure}[h]
\centering
\subfigure[Survey images]{
  \foreach \j in {1,2,...,8}  {
    \includegraphics[height=0.05\linewidth]{Appendix_image/Landscape/3/\j.jpg} 
  }
  }
\centering
\subfigure[Comparison of survey and predicting result by Alex base net]{
\includegraphics[scale=0.5]{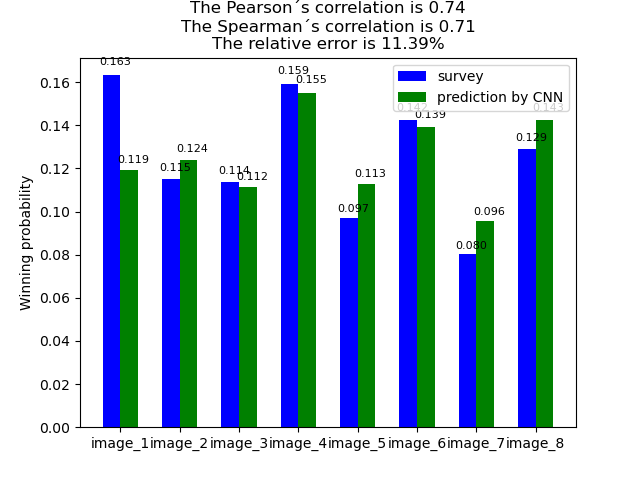}}
   \caption{Comparison of survey and prediction data of Alex base net for Blue Ridge Mountains.}
   \label{result_blue}
\end{figure}

The landscape images are used to train the CNN models in the above proposed method. Fig. \ref{result_blue} shows one group of results between the survey and the prediction. In the survey, Image 1 is ranked in the first place, followed by Image 4, 6 and 8, while in the CNN prediction model, Images 4, 8 and 6 are ranked in the first three places. Although there are some differences, the Spearman correlation coefficient between the predicting model and the survey result is 0.74.

\begin{figure}
  \centering
  \includegraphics[scale=0.5]{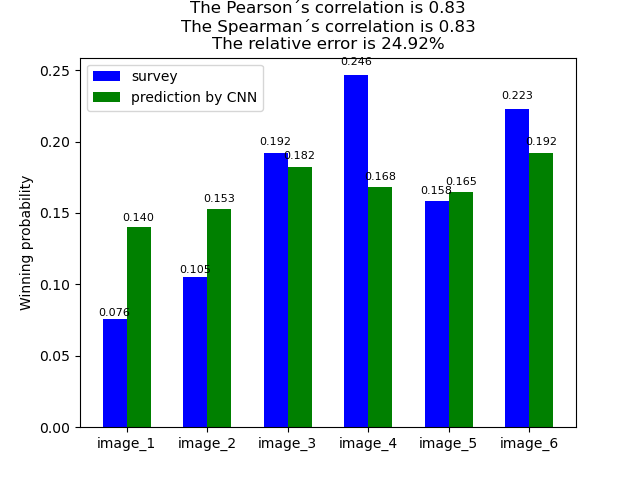}\\
  \caption{Comparison of survey and prediction data of Alex net for Yosmite.}
  \label{result_yosmite}
\end{figure}

Fig. \ref{result_yosmite} shows another example of results between the survey and the prediction of the Alex base net for Yosmite. In the survey, Image 6 has the highest score, while Image 1 is predicted in the first rank. Overall, the Pearson’s correlation and Spearman’s correlation coefficient are 0.83, which means that the prediction is significantly relevant for the survey result.

Table II shows the performance of the validation and the test dataset of different models for landscape images. As seen in this table, accuracy in pairs, PCC, SRCC and Average error usually show the same results, because higher accuracy means higher PCC and SRCC, and lower error at most times. Squeeze base net has the best performance with regards to validation data, with 61.06\% accuracy and 0.317 SRCC. VGG base net has the best performance regarding the test dataset, with 62.21\% accuracy and 0.320 SRCC.

In the table, LSiM base net shows negative correlation in the validation data and accuracy is lower than 50\%. This means that CNN fails to predict the result of HVS. However, the pretrained LSiM base net with 60.08\% accuracy in pairs is better than the one with initial values with 54.20\% accuracy in pairs. This means that the pretrained network indeed improves the proformance of CNN. For Alex base net and Squeeze base net, the result of the pretrained network is worse than that of the original network. 

Meanwhile, the quantitative evaluation by relative error of the different CNN models for landscape images is about 20\% for most models. The LSiM base model has the highest relative error in the test dataset with 34.18\%.
\begin{table*}[htbp]
\centering
\caption{Quantitative evaluation of the different CNN models for landscape images: Original base networks are compared with models pretrained by AVA dataset. Performance of different models are measured in term of accuracy (pairs), PCC, SRCC and average error. Bold and underlined values show the best result of CNN model for each dataset.}
\begin{tabular}{lllllllll}
\hline
\multirow{2}{*}{Models} & \multicolumn{4}{c}{Validation} & \multicolumn{4}{c}{Test}\\ 
\cmidrule(r){2-5} \cmidrule(r){6-9}
& Accuracy (pairs) & PCC    & SRCC   & Average error & Accuracy (pairs) & PCC   & SRCC  & Average error \\
\hline       
Alex base net                 & \underline{\textbf{61.06\%}}          & 0.244  & 0.309  & 17.13\%       & 59.24\%          & 0.379 & 0.256 & 27.56\%       \\
Alex base net (pretrained)    & 56.64\%          & \underline{\textbf{0.359}}  & 0.226  & 17.75\%       & 54.20\%          & 0.250 & 0.139 & 28.50\%       \\
VGG base net                  & 49.56\%          & 0.094  & 0.101  & 19.96\%       & \underline{\textbf{62.61\%}}          & 0.460 & \underline{\textbf{0.320}} & \underline{\textbf{26.06\%}}       \\
VGG base net (pretrained)     & 52.21\%          & 0.059  & 0.027  & 20.39\%       & 60.08\%          & 0.257 & 0.308 & 28.94\%       \\
Squeeze base net              & \underline{\textbf{61.06\%}}          & 0.290  & \underline{\textbf{0.317}}  & \underline{\textbf{17.00\%}}       & 60.08\%          & \underline{\textbf{0.483}} & 0.273 & 26.58\%       \\
Squeeze base net (pretrained) & 57.52\%          & 0.217  & 0.252  & 19.12\%       & 55.88\%          & 0.253 & 0.148 & 29.97\%       \\
LSiM base net                 & 41.59\%          & -0.180 & -0.207 & 23.35\%       & 54.20\%          & 0.007 & 0.141 & 34.18\%       \\
LSiM base net (pretrained)    & 49.56\%          & 0.036  & 0.012  & 19.60\%       & 60.08\%          & 0.251 & 0.221 & 28.07\%       \\
\hline 
\end{tabular}
\end{table*}

\subsection{Results of Portrait Images Assessment}
\begin{figure}[h]
\centering
\subfigure[Survey images]{
  \foreach \j in {1,2,...,5}  {
    \includegraphics[height=0.17\linewidth]{Appendix_image/Face/23/\j.jpg} 
  }
  }
\centering
\subfigure[Comparison of survey and predicting result]{
\includegraphics[scale=0.5]{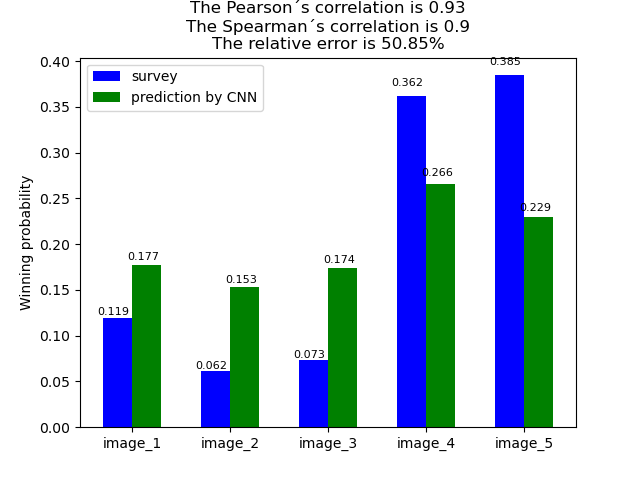}}
   \caption{Images and comparison results of group 23 between survey and prediction in different epochs during training process.}
   \label{result_man}
\end{figure}

Fig. \ref{result_man} compares one example of the survey results and the predicted results of the validation dataset for the middle-aged men. A high correlation between the survey and the predicted results can be observed with a PCC of 0.93 and a SRCC of 0.90.

Table II compares results of different CNN models. The Alex base net has the highest accuracy with 63.33\%, a PCC of 0.373 and a SRCC of 0.283 in the validation dataset, and a 75\% accuracy in the test dataset pairs. In second place, the VGG base net has a good performance with 61.67\% accuracy and a PCC of 0.238 in the validation dataset. The VGG base net's correlation in the test dataset is lower than the Alex base net's correlation, with a 58.33\% accuracy in pairs. The Squeeze base net has a higher accuracy with 66.67\% in the test dataset, but only 53.33\% in the validation dataset. The LSiM base net has the lowest correlation, both in the validation dataset and the test dataset.

In terms of quantitative evaluation, the Squeeze base net and the LSiM base net have the highest relative error in the validation data set, with 38.92\% and respectively 36.64\%. The relative error of the Alex base net and the VGG base net is comparably lower. The Alex base net has the lowest relative error about 27.97\% in the  validation dataset and 21.54\% in the test dataset. The results of the relative error show the same tendency as the correlation coefficient.

Overall, the prediction results of portrait images are better than those of landscape images with a higher correlation coefficients from about 60\% to 75\%. Meanwhile, the results of pretrained models are worse than original base net for the Alex and the VGG base net. For the Squeeze and the LSiM base net, the results of the validation dataset get better, but the results of the test dataset perform worse. These can be explained by the lack of portrait images in the AVA dataset. The pretrained features maps are not suitable for these portrait images. Therefore, pretrained models do not help improve performances in this case.
\begin{table*}[htbp]
\centering
\caption{Quantitative evaluation of the different CNN models for portrait images: Original base networks are compared with models pretrained by AVA dataset. Performance of different models are measured in term of accuracy (pairs), PCC, SRCC and average error. Bold and underlined values show the best result of CNN model for each dataset.}
\begin{tabular}{lllllllll}
\hline
\multirow{2}{*}{Models} & \multicolumn{4}{c}{Validation} & \multicolumn{4}{c}{Test}\\ 
\cmidrule(r){2-5} \cmidrule(r){6-9}
& Accuracy (pairs) & PCC    & SRCC   & Average error & Accuracy (pairs) & PCC   & SRCC  & Average error \\
\hline       
Alex base net                 & \underline{\textbf{63.33\%}}          & \underline{\textbf{0.373}}  & \underline{\textbf{0.283}}  & \underline{\textbf{27.97\%}}       & \underline{\textbf{75.00\%}}          & \underline{\textbf{0.632}} & \underline{\textbf{0.571}} & \underline{\textbf{19.10\%}} \\
Alex base net (pretrained)    & 55.00\%          & 0.244  & 0.067  & 31.06\%       & 70.00\%          & 0.464 & 0.462 & 28.71\%       \\
VGG base net                  & 61.67\%          & 0.238  & 0.267  & 36.17\%       & 58.33\%          & 0.238 & 0.201 & 25.94\%       \\
VGG base net (pretrained)     & 53.33\%          & 0.171  & 0.117  & 33.66\%       & 58.33\%          & 0.302 & 0.214 & 25.35\%       \\
Squeeze base net              & 53.33\%          & 0.259  & 0.050  & 38.92\%       & 66.67\%          & 0.480 & 0.386 & 20.30\%       \\
Squeeze base net (pretrained) & 63.33\%          & 0.232  & \underline{\textbf{0.283}}  & 41.40\%       & 60.00\%          & 0.240 & 0.227 & 26.57\%       \\
LSiM base net                 & 51.67\%          & 0.253  & 0.067  & 36.64\%       & 66.67\%          & 0.444 & 0.373 & 23.58\%       \\
LSiM base net (pretrained)    & 61.67\%          & 0.177  & \underline{\textbf{0.283}}  & 31.70\%       & 53.33\%          & 0.0l9 & 0.115 & 27.65\%       \\
\hline 
\end{tabular}
\end{table*}

\subsection{Results of Architecture Images Assessment}
\begin{figure}[h]
\centering
\subfigure[Survey images]{
  \foreach \j in {1,2,...,7}  {
    \includegraphics[height=0.075\linewidth]{Appendix_image/Buildings/10/\j.jpg} 
  }
  }
\centering
\subfigure[Survey results with exponential scores]{
\includegraphics[scale=0.5]{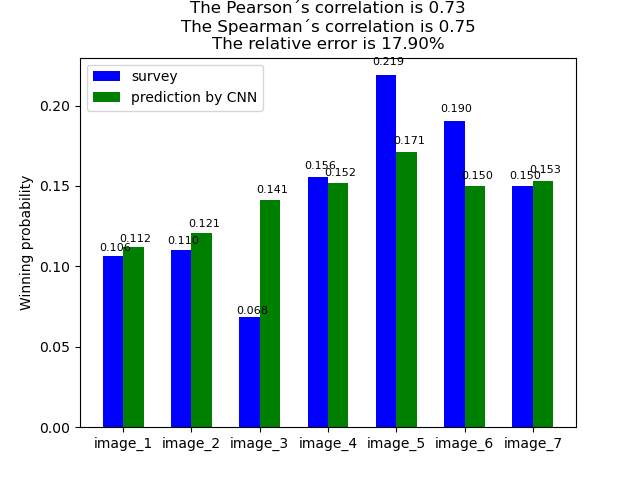}}
   \caption{Comparison of the survey and the prediction data for the Duomo Cathedral}
   \label{result_duomo}
\end{figure}

Fig. \ref{result_duomo} shows the result of the Duomo Cathedral comparison. In this group, the CNN model predicts the distribution of winning probabilities well. The PCC is 0.73 and the SRCC is 0.75, which implies a high correlation.

Table IV shows the average correlation coefficient of the validation and test dataset. The pretrained VGG base net has the best performance with a 64.76\% accuracy in pairs for the validation dataset and a 56.06\% accuracy for the test dataset. The other networks do not show a good correlation with negative coefficients. The Alex base net has a 53.33\% accuracy in the validation dataset and a 47.98\% accuracy in the test dataset. It can be seen that the performance of the validation dataset is better than the test dataset. This may be explained by the higher complexity of the test dataset. Other models do not show obvious correlations with the survey result. The quantitative evaluation shows that the pretrained VGG base net has the lowest relative error in the test dataset, which is in accordance with the results of both correlation coefficients. The pretrained squeeze base net has the highest error, almost 27.64\% in the validation dataset.

\begin{table*}[htbp]
\centering
\caption{Quantitative evaluation of the different CNN models for architecture images: Original base networks are compared with models pretrained by AVA dataset. Performance of different models are measured in term of accuracy (pairs), PCC, SRCC and average error. Bold and underlined values show the best result of CNN model for each dataset.}
\begin{tabular}{lllllllll}
\hline
\multirow{2}{*}{Models} & \multicolumn{4}{c}{Validation} & \multicolumn{4}{c}{Test}\\ 
\cmidrule(r){2-5} \cmidrule(r){6-9}
& Accuracy (pairs) & PCC    & SRCC   & Average error & Accuracy (pairs) & PCC   & SRCC  & Average error \\
\hline       
Alex base net                 & 53.33\%          & 0.152  & 0.124  & \underline{\textbf{20.34}}\%       & 47.98\%          & 0.078 & 0.038 & 22.86\%       \\
Alex base net (pretrained)    & 57.14\%          & 0.182  & 0.167  & 21.66\%       & 47.98\%          &-0.035 & -0.001& 22.33\%       \\
VGG base net                  & 49.52\%          &-0.114  &-0.015  & 25.05\%       & 47.98\%          &-0.081 &-0.081 & 25.35\%       \\
VGG base net (pretrained)     & \underline{\textbf{64.76\%}}          & \underline{\textbf{0.233}}  & \underline{\textbf{0.342}}  & 21.89\%       & 56.06\%          & \underline{\textbf{0.311}} & 0.152 & \underline{\textbf{20.59\%}}       \\
Squeeze base net              & 50.48\%          & 0.098  & 0.005  & 24.25\%       & 46.97\%          &-0.025 &-0.025 & 24.59\%       \\
Squeeze base net (pretrained) & 50.48\%          &-0.005  &-0.008  & 27.64\%       & 54.04\%          &-0.016 & 0.070 & 24.19\%       \\
LSiM base net                 & 52.38\%          &-0.089  & 0.031  & 23.65\%       & \underline{\textbf{58.08}}\%          & 0.198 & \underline{\textbf{0.249}} & 21.24\%       \\
LSiM base net (pretrained)    & 56.19\%          & 0.179  & 0.177  & 24.07\%       & 35.86\%          &-0.420 &-0.341 & 29.61\%       \\
\hline 
\end{tabular}
\end{table*}

Generally, the results of architecture images are worse than those of the landscape and portrait images. Many models show negative correlations and low accuracy. However, the pretrained networks perform better than the original base net as shown in Table IV. The AVA includes many architecture images and the pretraining by the AVA dataset helps to improve the performance of the CNN models. During the training process, frozen layers can also help to avoid overfitting. Pretraining in a large dataset and finetuning in a small dataset is one solution of this problem.

\section{Conclusion}
This work builds a CNN model to mimic HVS for image aesthetic evaluation. The motivation is to find a good way to automatically evaluate images due to the high costs of the survey. Therefore, training CNN models to replace HVS is a solution for evaluating images more efficiently.

To build models for image beauty assessment, such as landscape, portrait, and architecture images, surveys are conducted in a similar fashion as previously mentioned literature. The survey is implemented in pairs to make a comparable evaluation. The aesthetic scores are not directly given by the viewers, but instead are calculated by the Bradley-Terry models, which are used for the training of CNN models. Then four CNN models are compared with each other, and the results are evaluated by accuracy in pairs, correlation coefficients, and relative errors. As a result, portrait images have better results than landscape and architecture images. The pretrained networks by the AVA dataset do not improve the performance of portrait images, but it helps to predict the results of architecture images. Our model has a satisfactory  result with about 70\% accuracy in pairs.

The work aims to find a more scientific method to rate different images. Comparing the images in pairs is more accurate than simply letting the viewers score the images, because the direct comparison replace the need of scores. Future work would enlarge the training dataset, because the current image dataset is relatively small for a proper training of CNN models. The accuracy and correlation of prediction should also be improved to better mimic the HVS.


%

\appendices
\section{Architectures of Base networks}
In this work, different kinds of CNN models are used, which are introduced as follows. The results of different CNN models should be compared with each other to see the performance. Some CNN models are trained with initial weights, while others are pretrained and weights are updated according to the training data. The introduced base net serves as the basic architecture.
\subsection{Alex Base Net}
\begin{figure}
  \centering
  \includegraphics[scale=0.5]{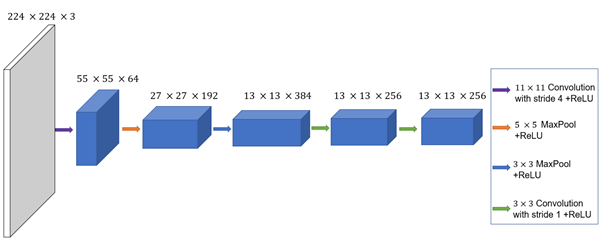}\\
  \caption{Architecture of Alex Base Net.}
  \label{alex}
\end{figure}

Alex Base Net is the convolutional part extracted from AlexNet \cite{krizhevsky2012imagenet}, which uses the pretrained Alex model to initialize the weights. The weights are updated by backpropagation during the training process. The pretrained network helps converge faster and extract specialized features. Fig. \ref{alex} shows the architecture of Alex base net.

The first convolutional layer filters the $224\times224\times3$ input image with 64 kernels of size $11\times11\times3$ and stride 4. The second layer uses MaxPool with 192 kernels of size $5\times5\times64$. The third, forth and fifth convolutional layers are connected to one another without any pooling. The third convolutional layer has 384 kernels of size $3\times3\times192$. The fourth layer has 256 kernels of size $3\times3\times 384$ and the fifth layer has 256 kernels of size $3\times3\times256$.

\subsection{LSiM Base Net}
There is not enough training data to update all the parameters of a big network. Therefore, in the LSiM base model \cite{kohl2020learning}, we reduce the number of feature maps by 50\% to fit the training data.

\begin{figure}
  \centering
  \includegraphics[scale=0.5]{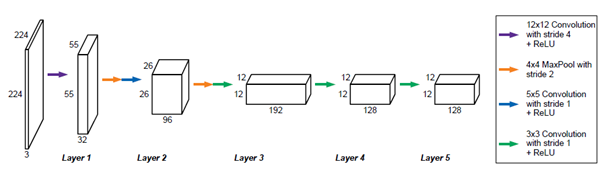}\\
  \caption{Architecture of LSiM Base Net: Proposed base network architecture extracted from AlexNet, where identical spatial dimensions for the feature maps are used, but it reduces the number of feature maps for each layer by 50\% to have fewer weights. \cite{kohl2020learning}}
  \label{lsim}
\end{figure}

Input image size of the base network is $224\times224$ pixels with three channels. In the first layer, $12\times12$ kernels,  are used with activation functions ReLU. Large kernels help to detect features with  larger scales. As the layer goes deeper, the size of the kernel gets smaller to extract finer features. In addition, MaxPool is used in the second and third layer to reduce the size of the feature map dramatically. In the forth and fifth layer, the size of the feature maps remains the same while the number of channels decrease, as shown in Fig. \ref{lsim}.
During the training process, these parameters are updated by backpropagation with loss functions.

\subsection{VGG Base Net}
\begin{figure}
  \centering
  \includegraphics[scale=0.5]{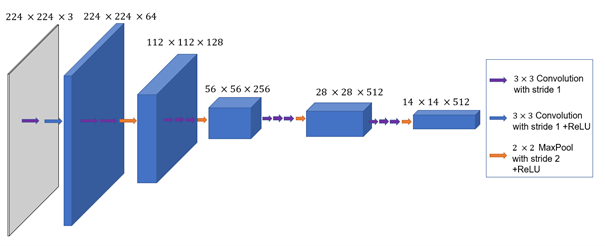}\\
  \caption{Architecture of VGG Base Net.}
  \label{vgg}
\end{figure}

The VGG base net is extracted from the pretrained VGG net \cite{simonyan2014very} without fully connected layers. From the second to the fifth layer, each part includes three convolution kernels and one $2\times2$ Max Pooling with stride 2, as shown in Fig. \ref{vgg}. It includes more convolution operations than the Alex base net, which means that it will be more time-consuming during the training process. CNN also has the ability to detect more complicated structures due to its large size.

\subsection{Squeeze Base Net}
\begin{figure}
  \centering
  \includegraphics[scale=0.5]{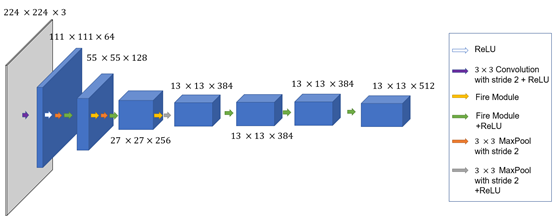}\\
  \caption{Architecture of Squeeze Base Net.}
  \label{squeeze}
\end{figure}

The pretrained Squeeze base net is inspired by the architecture of the SqueezeNet \cite{iandola2016squeezenet}. In the SqueezeNet, the fire module is introduced to reduce the size of the AlexNet. Fig. \ref{squeeze} shows the structure of the Squeeze base net, which is used in this work to train the model. The pretrained network will be loaded to initialize the parameters.


\section{Calculation Algorithm of Bradley-Terry Model}
This section introduces the calculation algorithm of the Bradley-Terry model, i.e. how the aesthetic scores are generated according to the number of votes. The most intuitive calculation algorithm is the optimization transfer method \cite{hunter2004mm}, which is based on the maximum likelihood scores for each sample.

Suppose we observe a number of pairings among $m$ samples and we wish to estimate the parameters $\gamma_i,\ldots,\gamma_m$ using maximum likelihood estimation. If outcomes of different pairings are assumed to be independent, the log-likelihood based on the Bradley-Terry model is
\begin{equation}
L(\gamma)=\sum_{i=1}^{m}\sum_{j=1}^{m}{[w_{ij}ln\gamma_i-w_{ij}ln(\gamma_i}+\gamma_j)]
\end{equation}

Denoting the number of votes where the participants chose image $i$ over image $j$ as $w_{ij}$ and assuming each vote is independent, we can represent the log likelihood for all pairs among all candidates by the defined score above:
\begin{equation}
L(s)=\sum_{i=1}^{m}{\sum_{j=1}^{m}{[w_{ij}s_i}-w_{ij}ln(e^{s_i}+e^{s_j}}])
\end{equation}

Based on the assumption that for any partition of the samples into sets A and B, at least one sample from A beats at least one sample from B at least once, we now describe an iterative algorithm to maximize $L(s)$. In start with an initial parameter vector $\gamma$, for $k = 1,2,\ldots,  $,the iteration process can be formulated as:
\begin{equation}
\begin{aligned}
{\gamma_i}^{(k+1)}&=W_i\left[\sum_{j\neq i}\frac{N_{ij}}{{\gamma_i}^{(k)}+{\gamma_j}^{(k)}}\right]^{-1}\\
&=W_i\left[\sum_{j\neq i}\frac{N_{ij}}{{e^{s_i}}^{(k)}+{e^{s_j}}^{(k)}}\right]^{-1}
\end{aligned}
\end{equation}
where $W_i$ denotes the number of wins by sample $i$ and $N_{ij}=w_{ij}+\ w_{ji}$ is the number of pairings between $i$ and $j$.

After counting the results, the vote number can be written in a list, which represents the votes of image $i$ against image $j$. As an example for this calculation algorithm, the matrix $w_m$ shows one of the survey results:

\begin{equation}
w_m=
\left[\begin{array}{ccccc}
0 & 21 & 29 & 16 & 22\\
33 & 0 & 36 & 19 & 32\\
25 & 18 & 0 & 15 & 28\\
38 & 35 & 39 & 0 & 34\\
32 & 22 & 26 & 20 & 0
\end{array}\right ] 
\end{equation}

As seen in the matrix, 21 participants think image 1 is more attractive than image 2, while 33 participants think image 2 is more attractive than image 1.

After adding the number of wins from each sample in each row, the wins vector $W$ can be calculated as:

\begin{equation}
W =
\left[\begin{array}{c}
88\\120\\86\\146\\100
\end{array}\right ] 
\end{equation}

After initializing parameter vector $\gamma^{1}$ as a unit vector, the final result can be calculated using equation (13) after 131 iterations:

\begin{equation}
\gamma^{132} =
\left[\begin{array}{c}
0.831\\1.358\\0.806\\2.051\\1.000
\end{array}\right ] 
\end{equation}

As it shows in vector $\gamma^{132}$, image 4 has the highest aesthetic score with 2.051, while image 3 has the lowest with 0.806.
\section{Images and Survey Results}

This section shows the images and survey results of different categories. Fig. 16 shows parts of images and survey results of landscape images. Fig. 16 (a) shows images, while Fig. 16 (b) presents scores of different images in one group. Fig. 17 shows data samples from portrait images, while Fig. 18 shows the data from architecture images.
\begin{figure*}[!h]
\centering
\foreach \i/\name[count=\cnt] in 
{8/Antarctic Patagonia,7/Arashiyama,8/Blue Ridge Mountains,8/Bryce Canyon,7/Cliffs of Moher,7/Elephant Island,7/Fiordland National Park,7/Godafoss} 
{ 
  \subfigure[Group \cnt: \name]{
  \foreach \j in {1,2,...,\i}  {
    \includegraphics[height=0.065\linewidth]{Appendix_image/Landscape/\cnt/\j.jpg} 
  }
  }
  \\
}
\centering
(a) Landscape images
\\
\foreach \i in {1,2,...,8}  {
    \includegraphics[width=0.22\linewidth]{Appendix_image/code_landscape/Result\i.png} 
}
  \\(b)Survey results
\caption{Data samples selected from landscape images and survey results: Images are listed for each group and survey results are showed in bar chart with scores of each.}
\end{figure*}

\begin{figure*}[!h]
\centering
\foreach \title/\name in 
{Little boy/1,Young boy/9,Young man/14,Man in middle age/21,Old man/28,Little girl/31,Young girl/37} 
{ 
  
  \foreach \j in {1,2,...,5}  {
    \includegraphics[width=0.15\linewidth]{Appendix_image/Face/\name/\j.jpg} 
  }
  \\
  Group \name: \title
}
\end{figure*}

\begin{figure*}[!h]
\centering
\foreach \title/\name/\i in 
{Young woman/52,Woman in middle age/57, Old woman/60} 
{ 
  
  \foreach \j in {1,2,...,5}  {
    \includegraphics[width=0.15\linewidth]{Appendix_image/Face/\name/\j.jpg} 
  }
  \\
  Group \name: \title
}
\\
\centering
 (a) Portrait images
\end{figure*}

\begin{figure*}[!h]
\foreach \i in {1,9,14,21,28,31,37,52,57,60}  {
    \includegraphics[width=0.22\linewidth]{Appendix_image/code_face/Result\i.png} 
}
  \centering
  \\(b) Survey results
\caption{Data samples selected from portrait images and survey results: Images are listed for each group and survey results are showed in bar chart with scores of each.}
\end{figure*}

\begin{figure*}[!h]
\centering
\foreach \i/\name[count=\cnt] in 
{8/Acropolis of Athens,7/Brandenburg Gate,9/Burj Khalifa,8/Chateau de Chenonceau,8/Chateau Frontenac,7/Cologne Cathedral} 
{ 
  \subfigure[Group \cnt: \name]{
  \foreach \j in {1,2,...,\i}  {
    \includegraphics[height=0.065\linewidth]{Appendix_image/Buildings/\cnt/\j.jpg} 
  }
  }
}
\\
\centering
(a) Architecture images
\\
\foreach \i in {1,2,...,6}  {
    \includegraphics[width=0.3\linewidth]{Appendix_image/code_building/Result\i.png} 
}
  \\(b)Survey results
\caption{Data samples selected from architecture images and survey results: Images are listed for each group and survey results are showed in bar chart with scores of each.}
\end{figure*}

\section*{Acknowledgment}

We would like to thank viewers online and research groups of Complex Fluids in Technical University Munich for filling out our questionnaires and rating the images according to their aesthetic values.

\ifCLASSOPTIONcaptionsoff
  \newpage
\fi



%
\bibliographystyle{plain}
\bibliography{bare_jrnl}




%

\begin{IEEEbiography}[{\includegraphics[width=1in,height=1.25in,clip,keepaspectratio]{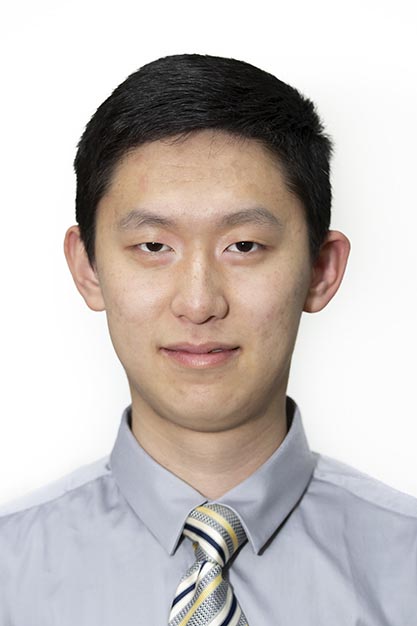}}]{Shiyu Li}

Shiyu Li received the B.S. degree from Beijing Institute of Technology in 2019. He is currently an M.S. student in the Department of Mechanical Engineering at the Technical University Munich. His research interests include image processing, computer vision and machine learning.

\end{IEEEbiography}

\begin{IEEEbiography}[{\includegraphics[width=1in,height=1.25in,clip,keepaspectratio]{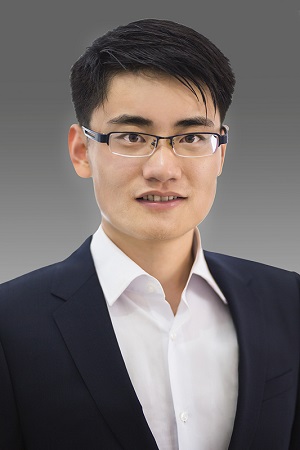}}]{Hao Ma}

Hao Ma received the B.S. degree from the Dalian University of Technology in 2015, and the M.S. degree from the National University of Defense Technology in 2017. He is currently a Ph.D. candidate in the School of Engineering and Design at the Technical University of Munich. His research interests include computer vision and machine learning.

\end{IEEEbiography}

\begin{IEEEbiography}[{\includegraphics[width=1in,height=1.25in,clip,keepaspectratio]{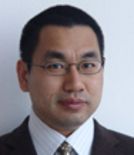}}]{Xiangyu Hu}

Xiangyu Hu received B. Sc. in Environmental Engineering, M. Sc. in Applied Physics, Ph.D. in Mechanical Engineering. He is now the Adjunct Teaching Professor, Reseach group leader on fluid mechanics at Technical University Munich. He has more than 20 years (since 1996) of experience on computational fluid and solid dynamics (CFSD) in research and developing advanced methods, mainly in three fields: compressible multi-material flows, high-order numerical algorithms and smoothed particle hydrodynamics.

He is current working on exploring the potential applications of these methods. The present research topics are on the mechanism of high-efficiency of fish swimming, the mechanism of slug production in pipe flow, the total-function heart simulation.

Other related topics are on image processing, including the technique on amplifying the image details which are dispersed by the lens, the halftone technique which is able to achieve fast generation of optimal dithered image.

\end{IEEEbiography}





\end{document}